\pdfoutput=1

\documentclass[11pt]{article}

\usepackage[final]{acl}

\usepackage{times}
\usepackage{latexsym}

\usepackage[T1]{fontenc}

\usepackage[utf8]{inputenc}

\usepackage{microtype}

\usepackage{inconsolata}

\usepackage[table]{xcolor}
\usepackage{microtype}
\usepackage{graphicx}
\usepackage{subfigure}
\usepackage{booktabs} %
\usepackage{amsmath}
\usepackage{array}
\usepackage{amssymb}
\usepackage{mathtools}
\usepackage{amsthm}
\usepackage{xspace}
\usepackage{enumitem}
\usepackage{wrapfig}
\usepackage{subcaption}
\usepackage{colortbl}
\usepackage{multirow}
\usepackage{tabulary}
\usepackage{graphicx}
\usepackage[capitalize,nameinlink,noabbrev]{cleveref}
\crefname{equation}{}{}
\newcommand{\ours}{ATS\xspace}

\usepackage{amsmath,amsfonts,bm}

\def\eqref#1{equation~\ref{#1}}

\def\1{\bm{1}}

\DeclareMathAlphabet{\mathsfit}{\encodingdefault}{\sfdefault}{m}{sl}
\SetMathAlphabet{\mathsfit}{bold}{\encodingdefault}{\sfdefault}{bx}{n}

\DeclareMathOperator*{\argmax}{arg\,max}

\usepackage{titlesec}
\titlespacing\subsection{0pt}{3pt plus 4pt minus 2pt}{0pt plus 2pt minus 2pt}
\titlespacing\subsubsection{0pt}{3pt plus 4pt minus 2pt}{0pt plus 2pt minus 2pt}

\definecolor{baselinecolor}{gray}{.9}
\newcommand{\baseline}[1]{\cellcolor{baselinecolor}{#1}}
\newlength\savewidth\newcommand\shline{\noalign{\global\savewidth\arrayrulewidth
  \global\arrayrulewidth 1pt}\hline\noalign{\global\arrayrulewidth\savewidth}}
\newcommand{\tablestyle}[2]{\setlength{\tabcolsep}{#1}\renewcommand{\arraystretch}{#2}\centering\footnotesize}

\newcolumntype{x}[1]{>{\centering\arraybackslash}p{#1pt}}
\newcolumntype{y}[1]{>{\raggedright\arraybackslash}p{#1pt}}
\newcolumntype{z}[1]{>{\raggedleft\arraybackslash}p{#1pt}}

\title{Calibrating Language Models with Adaptive Temperature Scaling}

\author{
  Johnathan Xie\thanks{Equal contribution.}, Annie S. Chen\footnotemark[1], Yoonho Lee, Eric Mitchell, Chelsea Finn \\
  Stanford University \\
\texttt{jwxie@stanford.edu, asc8@stanford.edu}
}

\begin{document}
\maketitle
\begin{abstract}

The effectiveness of large language models (LLMs) is not only measured by their ability to generate accurate outputs but also by their calibration—how well their confidence scores reflect the probability of their outputs being correct.
While unsupervised pre-training has been shown to yield LLMs with well-calibrated conditional probabilities, recent studies have shown that after fine-tuning with reinforcement learning from human feedback (RLHF), the calibration of these models degrades significantly.
In this work, we introduce Adaptive Temperature Scaling (\ours), %
a post-hoc calibration method that predicts a temperature scaling parameter for each token prediction. The predicted temperature values adapt based on token-level features and are fit over a standard supervised fine-tuning (SFT) dataset.
The adaptive nature of \ours addresses the varying degrees of calibration shift that can occur after RLHF fine-tuning.
\ours improves calibration by over 10-50\% across three downstream natural language evaluation benchmarks compared to prior calibration methods and does not impede performance improvements from RLHF.

\end{abstract}

\section{Introduction}
\label{sec:intro}

Large language models (LLMs) have become a cornerstone of modern artificial intelligence, offering impressive capabilities in natural language processing tasks. 
However, the reliability of LLMs is intertwined with their ability to generate confidence scores that accurately reflect the likelihood of their outputs being correct. 
This calibration, aligning a model's confidence with its accuracy, is essential, especially when LLMs are deployed in real-world scenarios where decisions based on incorrect outputs can have significant consequences.

While unsupervised pre-training methods have shown success in producing well-calibrated LLMs, a challenge arises when these models undergo fine-tuning through reinforcement learning from human feedback (RLHF). 
While RLHF fine-tuning is effective in enhancing model performance on specific tasks and aligning outputs with human preferences, recent studies indicate a notable degradation in the calibration of LLMs post-RLHF fine-tuning~\citep{achiam2023gpt4,tian2023just,kadavath2022language}. This degradation compromises the model’s ability to provide reliable confidence scores, an issue that becomes critical when these models are applied to tasks requiring high levels of trust and accuracy.
An important question arises: how can we maintain the performance gains achieved through RLHF fine-tuning while ensuring that the model's confidence scores remain reliable?

To address this challenge, our work introduces Adaptive Temperature Scaling (\ours), a post-hoc calibration technique that predicts a temperature scaling parameter for each token prediction based on a language model's hidden features. 
Basic temperature scaling is a widely-used calibration method that applies a single temperature parameter across all outputs of a model. 
This technique, while effective in some contexts, assumes uniform calibration needs across all inputs, which is often not the case for complex models like LLMs. 
\ours, in contrast, predicts a unique temperature scaling parameter for each set of token predictions. 
This input-specific approach allows \ours to refine the calibration process, addressing the varying degrees of calibration shift that can occur after RLHF fine-tuning. 
For instance, certain inputs or topics might be more susceptible to miscalibration post-RLHF, and \ours can adaptively adjust the scaling for these instances more aggressively than for others where the model's confidence remains relatively well-aligned with its accuracy.
Importantly, our approach reduces the need for task-specific calibration, which may be difficult to achieve in many cases, given the wide variety of downstream tasks that LLMs may be used for.

We conduct experiments on MMLU, TriviaQA, and TruthfulQA to evaluate the effectiveness of \ours in improving the calibration of LLMs following RLHF fine-tuning. 
Our findings demonstrate that \ours improves the calibration of post-RLHF LLMs by 10-50\% on average, while having no effect on model performance.

\section{Related Work}
\label{sec:related_work}

Recent literature has extensively discussed the challenges of maintaining calibration in LLMs, particularly highlighting the degradation in calibration post-RLHF~\citep{lin2022teaching,park2022calibration,kadavath2022language,xiao2022uncertainty,kuhn2023semantic}. 
The concept of verbalized confidence has been explored as a way to counteract this degradation~\citep{xiong2023can,tian2023just}, and dialogue models have been shown to express uncertainty in a well-calibrated manner~\citep{mielke2022reducing,zhou2023navigating}.
Compared to works on improving sentence level calibration given token-level probabilities~\cite{kuhn2023semantic,tian2023just}, our work aims to directly improve the calibration of token-level probabilities.

The calibration of neural networks has been a topic of significant interest, with foundational concepts such as proper scoring rules~\citep{gneiting2007probabilistic} laying the groundwork. 
Model mismatch and distribution shift often degrade calibration, commonly quantified with common metrics including Expected Calibration Error (ECE)~\citep{naeini2015obtaining} and Brier score~\citep{brier1950verification}.
Modern neural networks have been found to exhibit overconfidence~\citep{guo2017calibration,thulasidasan2019mixup,wen2020combining}, especially in the context of image classification~\citep{geirhos2018imagenet,taori2020measuring,wen2020combining,hendrycks2021many}. 

Various methods have been proposed for calibrating neural networks, including temperature scaling~\citep{guo2017calibration}, Platt scaling~\citep{platt1999probabilistic,niculescu2005predicting}, label smoothing~\citep{muller2019does}, scaling binning~\citep{kumar2019verified,zhang2023study}, and more sophisticated approaches~\citep{hendrycks2018deep,katz2022training,choi2023conservative,jiang2023calibrating}. 
While these methods offer strategies for improving model calibration, our approach uniquely adapts the temperature scaling parameter for each token prediction based on its hidden features, tailoring the method to the problem of language modeling.

\section{Background and Problem Setting}
\label{sec:setting}
We consider access to a conversation SFT dataset of $\mathcal{D} = \{(x, y)\}$ with vocabulary $V$ where $x \in V^{l_{x}}$, denotes the instruction, each with sequence length $l_{x}$, and $y \in V^{l_y}$ is the corresponding response with sequence length $l_y$. 
We wish to calibrate language model $\pi(y|x)$. While we do not make any assumptions about the training process of $\pi$, we find our calibration method is most useful for language models following an RLHF process where token-level calibration is often significantly degraded compared to base language models which are generally well calibrated~\cite{achiam2023gpt4}.

For a given sample $(x, y)$, we generate a set of unnormalized logits $\hat{z} = \pi(x) \in \mathbb{R}^{l_x + l_y \times |V|}$ where each $\hat{z}_i$ defines the unnormalized logits for the $i + 1$-th token and $|V|$ is the vocabulary size. Prior methods~\cite{guo2017calibration,platt1999probabilistic} propose various scaling methods for calibrating models by transforming logits. 
In matrix scaling, a calibration head is used to produce calibrated logits $\hat{q} = W\hat{z} + b$ where $W, b$ are learnable parameters. In the case of language modeling where $|V|$ is large, learning a full transform matrix becomes computationally infeasible, so we compare to vector scaling, where $W$ is constrained to a diagonal matrix. Temperature scaling is the case when $W$ is constrained further to a scalar matrix and $b$ to the zero-vector. To learn these parameters, these methods minimize the cross-entropy over the SFT dataset calculated over response tokens.

\section{Adaptive Temperature Scaling}
\paragraph{Architecture.}
Temperature scaling, while effective in classification settings, struggles to adapt logits well in language modeling as the confidence scores that are most important (such as those that contain actual answers or facts) account for only a small portion of natural language sequences. Therefore, optimizing a single temperature parameter often results in post-RLHF language models still being overconfident post scaling. Additionally, language model miscalibration largely varies based on the type of token being predicted following RLHF.
Matrix and vector scaling can in theory perform adaptive confidence prediction by using logits as features; however, they are prone to overfitting, as we find in~\cref{sec:experiments}.

To balance regularization with modeling capacity in our calibration head, we instead propose to use a head architecture that predicts a singular temperature for every token prediction. For an input pair $(x, y)$, we first produce input-dependent features $\hat{h} \in \mathbb{R}^{l_x + l_y, h}$ using the language model $\pi$.

We then learn a calibration head to produce a temperature vector $c_\theta(\hat{h}) = \tau \in \mathbb{R}^{l_x + l_y}$. We exponentiate $\tau$ to ensure positive values then transform logits to yield calibrated logits $\hat{q} = \hat{z} \circ e^\tau$. In practice, we find that directly using the logits $\hat{z}$ as features can be inefficient (with a large vocabulary size) and also less effective compared to hidden states. Therefore, we use the last hidden state of the language model $\pi$ as the features for predicting $\tau$. With this architecture formulation, we retain the ability to predict confidences adaptively depending on the context, while also never changing the ranking for the possible next token given specific context, as each set of token logits are scaled by only a single value.

\paragraph{Loss function.}
To improve the process of calibration, we take inspiration from selective classification works~\citep{choi2023conservative} and use a loss function which adapts targets depending on the correctness of the original language model. For a logit, label pair $\hat{q} \in \mathbb{R}^v$, $y \in V$, and weighting hyperparameter $\alpha \in [0, 1]$ we optimize the  following loss function $\ell$:
\begin{equation}
    \ell(\hat{q}, y) = 
    \begin{cases} 
      -(1 - \alpha)\log\left(
      \sigma_{SM}(\hat{q})_y
      \right) & \argmax \hat{q} = y \\
      -\frac{\alpha}{|V|}\sum_{i=1}^{|V|}\log(\sigma_{SM}(\hat{q}))_i & \argmax \hat{q} \neq y
   \end{cases}
\end{equation}
This loss function uses a uniform distribution as the target when the model is incorrect and a standard one-hot cross-entropy when the model is correct. 

\section{Experiments}
\label{sec:experiments}

In this section, we aim to evaluate our proposed method on multiple benchmarks to demonstrate its effectiveness in improving calibration of LLMs fine-tuned with RLHF. 
We compare our method to no calibration as well as existing temperature scaling methods. 
Additionally, we ablate the main components of our method including the loss function, loss weighting, and head architecture.

\begin{table*}[tbp]

\begin{center}
\resizebox{0.9\textwidth}{!}{
    \begin{tabular}{llccccccccc}
        \toprule
        \multirow{2}{*}{Model} & \multirow{2}{*}{Calibration} & \multicolumn{3}{c}{MMLU} & \multicolumn{3}{c}{TriviaQA} & \multicolumn{3}{c}{TruthfulQA}  \\
        \cmidrule(lr){3-5} \cmidrule(lr){6-8} \cmidrule(lr){9-11}
        & & Acc & ECE & BS & Acc & ECE & BS & Acc & ECE & BS \\
        Llama-2-7b-Chat~\citep{touvron2023llama2} & None & 0.474 & 0.298 & 0.313 & 0.592 & 0.221 & 0.239 & 0.322 & 0.507 & 0.480 \\
        & Temperature & 0.474 & 0.270 & 0.295 & 0.592 & 0.187 & 0.224 & 0.322 & 0.492 & 0.463\\
        & Vector Scaling & 0.474 & 0.324 & 0.333 & 0.592 & 0.211 & 0.234 & 0.322 & 0.499 & 0.471\\
        & Scaling Binning & 0.474 & 0.296 & 0.312 & 0.592 & 0.222 & 0.239 & 0.322 & 0.544 & 0.504 \\
        & ATS (Ours) & 0.474 & \textbf{0.125} & \textbf{0.227} & 0.592 & \textbf{0.069} & \textbf{0.217} & 0.322 & \textbf{0.197} & \textbf{0.264} \\
        \midrule 
        Qwen-7b-Chat~\citep{bai2023qwen} & None & 0.571 & 0.141 & 0.215 & 0.495 & 0.272 & 0.311 & 0.230 & 0.372 & 0.304 \\
        & Temperature & 0.571 & 0.093 & 0.215 & 0.495 & 0.269 & 0.308 & 0.230 & 0.313 & 0.262 \\
        & Vector Scaling & 0.571 & 0.144 & 0.218 & 0.495 & \textbf{0.252} & 0.308 & 0.230 & 0.369 & 0.302 \\
        & Scaling Binning & 0.571 & 0.132 & 0.324 & 0.495 & 0.320 & 0.431 & 0.230 & 0.385 & 0.308 \\
        & ATS (Ours) & 0.571 & \textbf{0.050} & \textbf{0.190} & 0.495 & 0.254 & \textbf{0.303} & 0.230 & \textbf{0.165} & \textbf{0.188} \\
        \midrule
        Llama-2-13b-Chat~\citep{touvron2023llama2} & None & 0.532 & 0.228 & 0.262 & 0.679 & 0.150 & 0.200 & 0.368 & 0.484 & 0.461 \\
        & Temperature & 0.532 & 0.175 & 0.235 & 0.679 & 0.065 & \textbf{0.185} & 0.368 & 0.443 & 0.418 \\
        & Vector Scaling & 0.532 & 0.246 & 0.283 & 0.679 & 0.120 & 0.191 & 0.368 & 0.378 & 0.371 \\
        & Scaling Binning & 0.532 & 0.227 & 0.260 & 0.679 & 0.150 & 0.199 & 0.368 & 0.494 & 0.466 \\
        & ATS (Ours) & 0.532 & \textbf{0.092} & \textbf{0.211} & 0.679 & \textbf{0.061} & 0.200 & 0.368 & \textbf{0.192} & \textbf{0.267} \\
        \midrule
    \end{tabular}
    }
\end{center}
\vspace{-1em}
\caption{\textbf{Model Calibration Comparison}. We find that \ours yields significant improvements over other calibration methods for both LLama-2-7b-Chat and Qwen-7b-Chat.
}
\label{tab:main}
\end{table*}

\begin{table*}[t]
\vspace{-.2em}
\vspace{-2mm}
\centering

\subfigure[
\textbf{Smoothing type}. Selective smoothing outperforms cross-entropy (no smoothing) and label smoothing (full smoothing).
\label{tab:loss_type}
]{
\centering
\begin{minipage}{0.29\linewidth}{
\begin{center}
\tablestyle{4pt}{1.05}
\begin{tabular}{y{60}x{24}x{24}}
loss & ECE & BS \\
\shline
no smoothing & 0.226 & 0.269 \\
full smoothing & 0.149 & 0.236 \\
selective & \baseline{\textbf{0.125}} & \baseline{\textbf{0.227}} \\
& & \\
& & \\
& & \\
\end{tabular}
\end{center}
}\end{minipage}
}
\hfill
\vspace{-3mm}
\subfigure[
\textbf{Loss weighting}. A high smooth loss weight is necessary to correct for language model overconfidence.
\label{tab:loss_weighting}
]{
\begin{minipage}{0.29\linewidth}{
\begin{center}
\tablestyle{4pt}{1.05}
\begin{tabular}{x{18}x{24}x{24}}
$\alpha$ & ECE & BS \\
\shline
0.1 & 0.197 & 0.254 \\
0.2 & 0.172 & 0.243 \\
0.3 & 0.151 & 0.236 \\
0.4 & 0.134 & 0.231 \\
0.5 & \baseline{0.125} & \baseline{0.227} \\
0.6 & \textbf{0.113} & \textbf{0.224} \\
\end{tabular}
\end{center}
}
\end{minipage}
}
\hfill
\subfigure[
\textbf{Head architecture}. We find that using a Transformer head in the same configuration as LLaMa-2-7b-Chat performs best.
\label{tab:head_architecture}
]{
\begin{minipage}{0.29\linewidth}{\begin{center}
\tablestyle{1pt}{1.05}
\begin{tabular}{y{56}x{24}x{24}}
head & ECE & BS \\
\shline
linear & 0.140 & 0.233 \\
mlp &  0.132 & 0.230 \\
transformer & \baseline{\textbf{0.125}} & \baseline{\textbf{0.227}} \\
& & \\
& & \\
& & \\
\end{tabular}
\end{center}}
\end{minipage}
}
\\
\vspace{.3em}
\label{tab:ablations} 
\vspace{-1em}
\end{table*}

\paragraph{Evaluation Setting.}
We evaluate using two 7B parameter post-RLHF models LLama-2-Chat-7b~\cite{touvron2023llama2} and Qwen-Chat-7b. As the calibration dataset, we use the Alpaca GPT-4~\cite{peng2023alpacagpt4} instruction tuning dataset, which contains a diverse set of instructions with high quality answers. We then evaluate model calibration on three downstream tasks.

We perform multiple choice evaluation on the MMLU~\cite{hendrycks2020mmlu} by aggregating statistics across the entire dataset. Specifically we concatenate the confidences and correctness labels from all subjects, then calculate the calibration metrics. We also evaluate on two free response datasets, TriviaQA~\cite{joshi2017triviaqa} and TruthfulQA~\cite{lin2021truthfulqa}. 

\paragraph{Metrics.}
In multiple choice inference, we have a set of tokens ids $O$ which represent the valid options for a multiple choice answer, so the confidence scores are $p = \sigma_{SM}(\hat{q}_{l_x,j \in O})$ where $\sigma_{SM}$ denotes the softmax function.
To calculate confidences over a long sequence of response tokens for an input $x$, we sample a generation $\hat{y}$ of length $l_{\hat{y}}$ from the original language model then concatenate to the instruction to form $\hat{z}$ and $\hat{q}$ following calibration. Then, we calculate an average over transition probabilities on the response tokens.
We use the Expected Calibration Error (ECE)~\cite{guo2017calibration} and Brier score~\cite{brier1950verification} to evaluate calibration. We also report accuracy but each method does not significantly affect accuracy.

\paragraph{Baselines.}
We compare our method to the post-RLHF model without calibration, temperature scaling, vector scaling, and scaling binning~\citep{kumar2019verified,zhang2023study}.
We do not evaluate matrix scaling as the full matrix becomes computationally infeasible for large vocabulary sizes, as the projection matrix requires the square of the vocabulary size parameters.

\subsection{Results}

We report the results of our method compared to the baselines in Table~\ref{tab:main}. Overall, we find that our method improves calibration by 10-50\% across the three benchmarks in terms of ECE and Brier Score compared to the next best method for both LLama-2-7b-Chat and Qwen-7b-Chat.
More specifically, for Llama-7b-Chat, applying \ours achieved the lowest ECE and BS across all downstream benchmarks, showing how adjusting the temperature scaling parameter for each token prediction can significantly improve calibration.
Qwen-7b-Chat also saw a significant improvement in calibration, although in the case of TriviaQA, \ours actually makes Qwen-7b-Chat slightly underconfident compared to vector scaling.
Importantly, the calibration dataset used for training \ours, Alpaca GPT-4, is unrelated to the downstream tasks evaluated on, which suggests that the method does not overfit to the calibration data but rather captures underlying predictive uncertainty principles applicable across various tasks.

\subsection{Ablation Studies}
To analyze our method, we ablate the main components: loss objective, loss weight, and head architecture, measuring calibration metrics on MMLU.

\paragraph{Loss objective.}
We compare different loss objectives, standard cross-entropy, cross-entropy with label smoothing, and selective smoothing (ours) in Table~\ref{tab:loss_type}. For label smoothing we performed a sweep and found a smoothing value of 0.3 to be optimal. We find that selective smoothing outperforms both the typical cross-entropy loss and label smoothing. One possible explanation for cross-entropy and standard label smoothing being less effective is that learning adaptive temperature values with a cross-entropy loss can actually cause the model to increase confidence when the model is incorrect. In comparison, by using a uniform distribution target for incorrect predictions, this will never happen.

\paragraph{Loss weight.}
We perform a sweep of smooth loss weight in Table~\ref{tab:loss_weighting}. While increasing the loss weight to 0.6 (compared to 0.5) benefits MMLU calibration, in practice we found this higher loss weight began to perform worse for TriviaQA, and we did not sweep higher values as the model begins to become underconfident.

\paragraph{Head architecture.}
In Table~\ref{tab:head_architecture}, we ablate the choice of head architecture. We find that a causal transformer layer identical to those used in the LLama-2-7b-chat model performs best. Given that the inference cost of a single additional layer is relatively negligible, using a full transformer layer is generally best for calibration performance as it can aggregate hidden state values from prior tokens for the specific task of predicting calibration.

\section{Conclusion}
In this paper, we introduce Adaptive Temperature Scaling (\ours), a novel calibration technique for large language models (LLMs) fine-tuned with reinforcement learning from human feedback (RLHF), offering a significant improvement in model calibration without compromising post-RLHF performance. 
By adapting the temperature scaling parameter based on token-level features of each input, \ours addresses the diverse calibration needs of LLMs. 
Our results across multiple benchmarks confirm our approach's efficacy in maintaining calibration post-RLHF. 
\section{Limitations}
While \ours offers a significant improvement in model calibration without compromising post-RLHF performance by adapting the temperature scaling parameter based on token-level features of each input, limitations remain. 
In particular, we do not test how \ours interacts with different sentence-level confidence methods such as semantic uncertainty. 
These limitations underscore the need for ongoing research to refine calibration techniques and incorporate a more nuanced understanding of  uncertainty to develop methods that allow models to express confidence in a manner that aligns with natural language.

\section*{Acknowledgements}
We thank anonymous reviewers for their helpful feedback.
This work was supported by an NSF graduate fellowship, Microsoft Azure, Apple, Juniper, and ONR grant N00014-20-1-2675.

\bibliography{main}

\clearpage
\newpage
\newpage
\appendix
\begin{figure*}[t!]
\centering
   \includegraphics[width=1.0\linewidth]{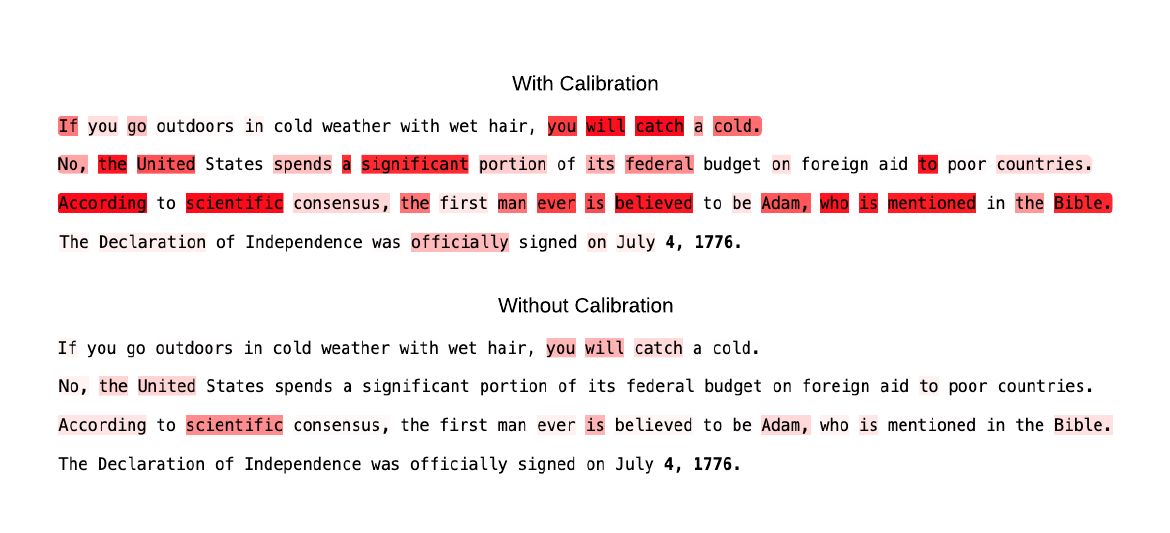}
   \vspace{-1cm}
   \caption{
   \textbf{Calibration Visualization.} 
   We visualize confidence calibration samples, comparing token-wise confidences before and after calibration. The less confident a token is, the more red we highlight the background. Additionally, we average the confidences of tokens to form full words in order to create a more interpretable visualization.
   }
   \label{fig:cal_vis}
   \vspace{-2em}
\end{figure*}

\section{Confidence Visualizations}
In Figure~\ref{fig:cal_vis}, we compare confidence calibration on TruthfulQA dataset samples. We compare the Llama-2-7b-chat model without any calibration to after calibration with our method. Our method is able to cause the language model to become significantly less confident on tokens containing inaccuracies.
\section{Hyperparameters}
\begin{table}[h]
    \centering
    \fontsize{8pt}{10pt}\selectfont
    \begin{tabular}{y{80}|x{60}}
    config & value \\
    \hline
    optimizer & AdamW \\
    optimizer betas & $\beta_1, \beta_2{=}0.9, 0.999$ \\
    weight decay & 0.0 \\
    learning rate & $5e-5$ \\
    learning rate schedule & cosine decay \\
    epochs & 2 \\
    batch size & 8 \\
    \end{tabular}%
    \caption{Calibration training hyperparameters.}
\label{tab:calibration_hyperparameters}
\end{table}

In Table~\ref{tab:calibration_hyperparameters} we list the main hyperparameters used for training calibration methods over Alpaca GPT-4.

\section{Discussion on Computational Costs}
\ours involves fine-tuning language models, and it takes approximately 6 L40 GPU hours (6 hours on a single L40 GPU) to fine-tune Llama-7b for 2 epochs over Alpaca GPT-4 English. In terms of additional inference cost, the forward pass is 1.04 seconds for the base model and 1.12 seconds when applying our method. We find that the total additional computational cost of our method is relatively small, and the additional forward pass cost can likely be further reduced with better optimized code as the cost is only a single additional transformer layer or 1/32th the cost of a full Llama-7b model. 

\section{Reliability Diagrams}
\begin{figure*}[htbp]
    \centering
    \begin{minipage}{0.45\textwidth}
        \centering
        \subfigure[Uncalibrated Llama-2-7b-Chat MMLU reliability diagram]{
            \includegraphics[width=\textwidth]{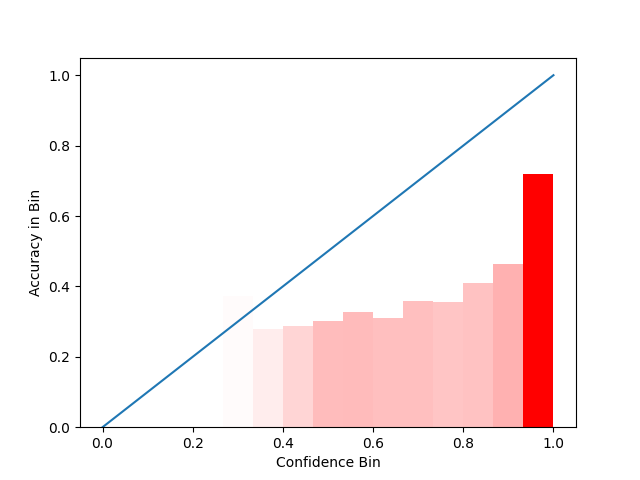}
        }
    \end{minipage}
    \hfill
    \begin{minipage}{0.45\textwidth}
        \centering
        \subfigure[Calibrated Llama-2-7b-Chat MMLU reliability diagram]{
            \includegraphics[width=\textwidth]{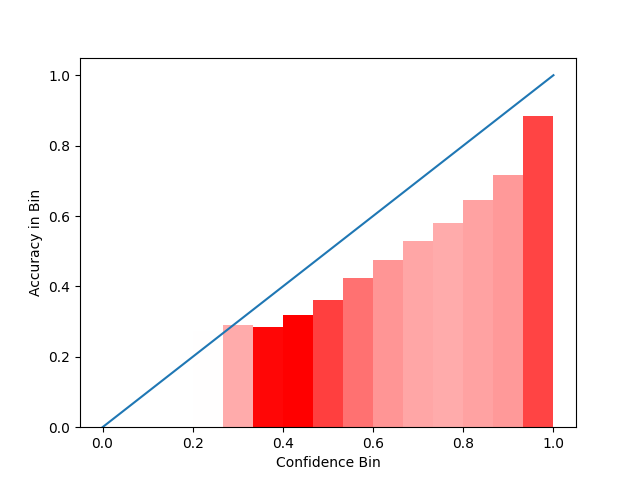}
        }
    \end{minipage}
    \vskip\baselineskip
    \begin{minipage}{0.45\textwidth}
        \centering
        \subfigure[Uncalibrated Llama-2-7b-Chat TriviaQA reliability diagram]{
            \includegraphics[width=\textwidth]{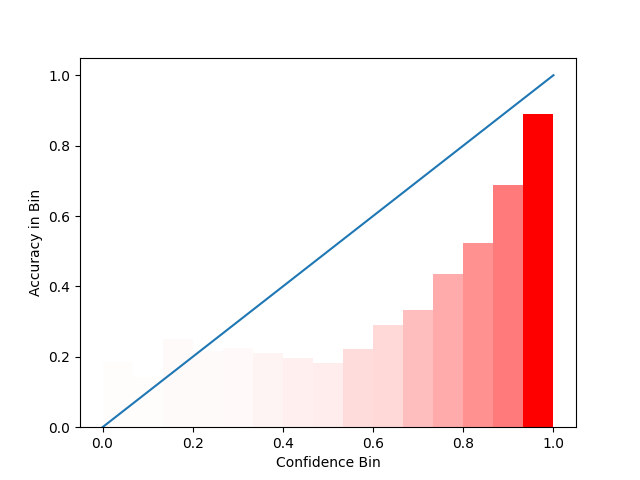}
        }
    \end{minipage}
    \hfill
    \begin{minipage}{0.45\textwidth}
        \centering
        \subfigure[Calibrated Llama-2-7b-Chat TriviaQA reliability diagram]{
            \includegraphics[width=\textwidth]{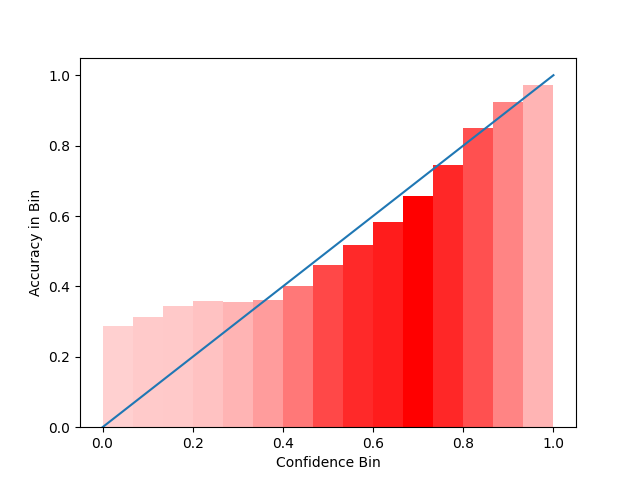}
        }
    \end{minipage}
    \vskip\baselineskip
    \begin{minipage}{0.45\textwidth}
        \centering
        \subfigure[Uncalibrated Llama-2-7b-Chat TruthfulQA reliability diagram]{
            \includegraphics[width=\textwidth]{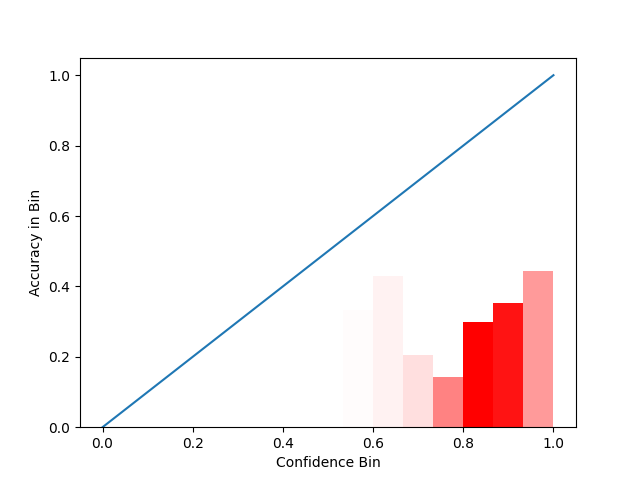}
        }
    \end{minipage}
    \hfill
    \begin{minipage}{0.45\textwidth}
        \centering
        \subfigure[Calibrated Llama-2-7b-Chat TruthfulQA reliability diagram]{
            \includegraphics[width=\textwidth]{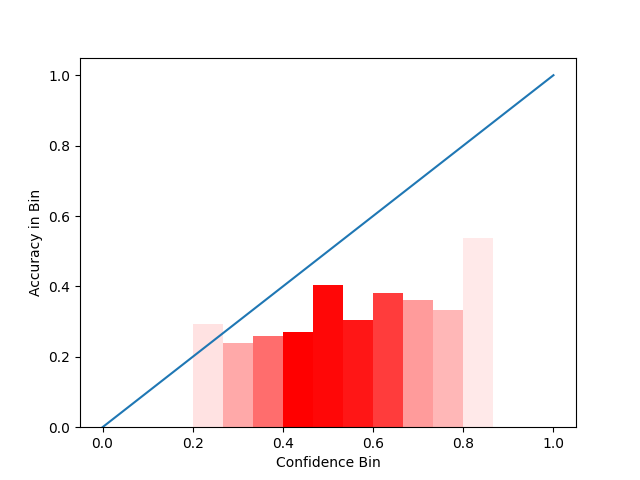}
        }
    \end{minipage}
    \caption{Llama-2-7b-Chat reliability diagrams.}
    \label{fig:llama7b_reliability_diagrams}
\end{figure*}

\begin{figure*}[htbp]
    \centering
    \begin{minipage}{0.45\textwidth}
        \centering
        \subfigure[Uncalibrated Qwen-7b-Chat MMLU reliability diagram]{
            \includegraphics[width=\textwidth]{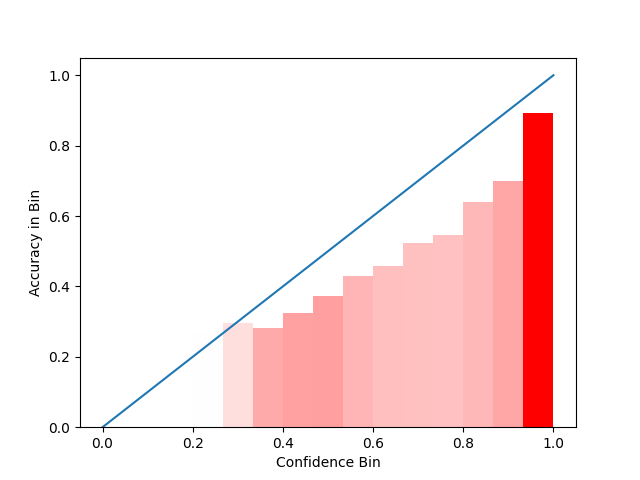}
        }
    \end{minipage}
    \hfill
    \begin{minipage}{0.45\textwidth}
        \centering
        \subfigure[Calibrated Qwen-7b-Chat MMLU reliability diagram]{
            \includegraphics[width=\textwidth]{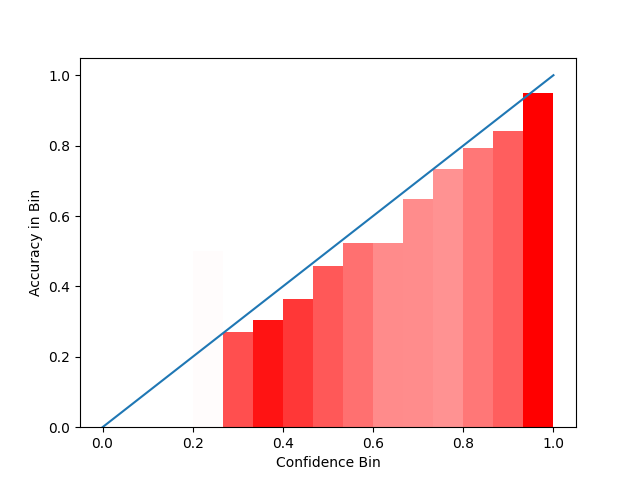}
        }
    \end{minipage}
    \vskip\baselineskip
    \begin{minipage}{0.45\textwidth}
        \centering
        \subfigure[Uncalibrated Qwen-7b-Chat TriviaQA reliability diagram]{
            \includegraphics[width=\textwidth]{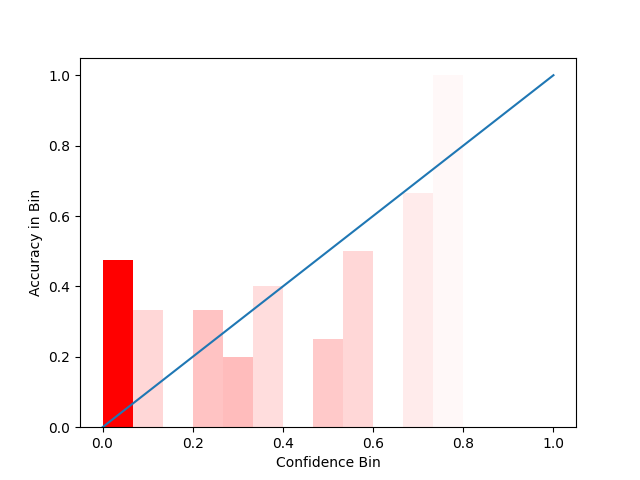}
        }
    \end{minipage}
    \hfill
    \begin{minipage}{0.45\textwidth}
        \centering
        \subfigure[Calibrated Qwen-7b-Chat TriviaQA reliability diagram]{
            \includegraphics[width=\textwidth]{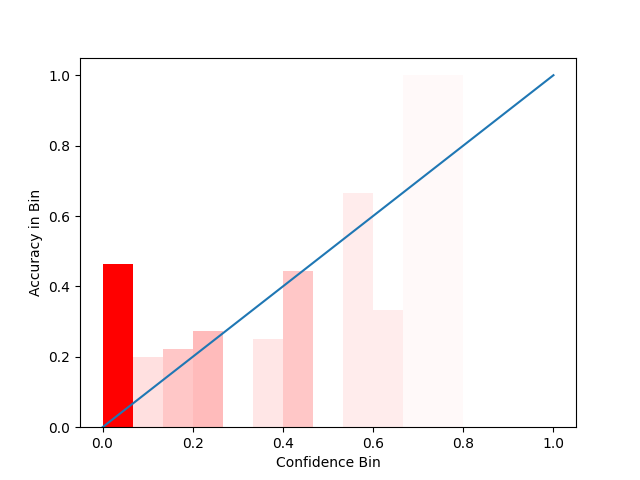}
        }
    \end{minipage}
    \vskip\baselineskip
    \begin{minipage}{0.45\textwidth}
        \centering
        \subfigure[Uncalibrated Qwen-7b-Chat TruthfulQA reliability diagram]{
            \includegraphics[width=\textwidth]{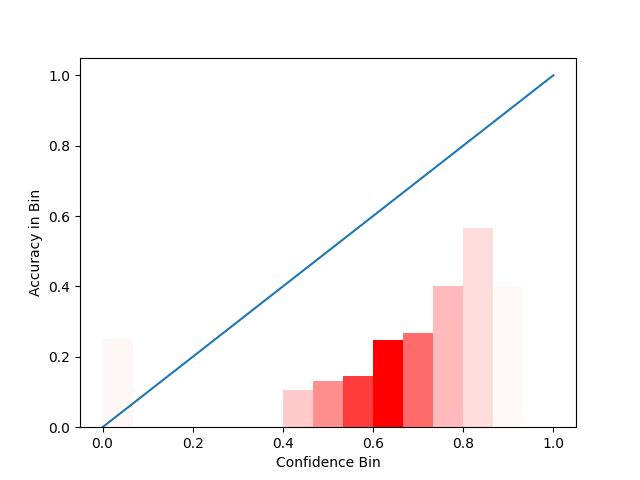}
        }
    \end{minipage}
    \hfill
    \begin{minipage}{0.45\textwidth}
        \centering
        \subfigure[Calibrated Qwen-7b-Chat TruthfulQA reliability diagram]{
            \includegraphics[width=\textwidth]{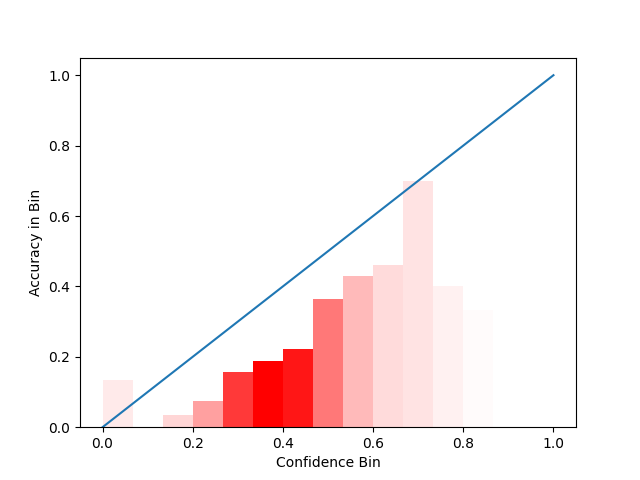}
        }
    \end{minipage}
    \caption{Qwen-7b-Chat reliability diagrams.}
    \label{fig:qwen_reliability_diagrams}
\end{figure*}

\begin{figure*}[htbp]
    \centering
    \begin{minipage}{0.45\textwidth}
        \centering
        \subfigure[Uncalibrated Llama-2-13b-Chat MMLU reliability diagram]{
            \includegraphics[width=\textwidth]{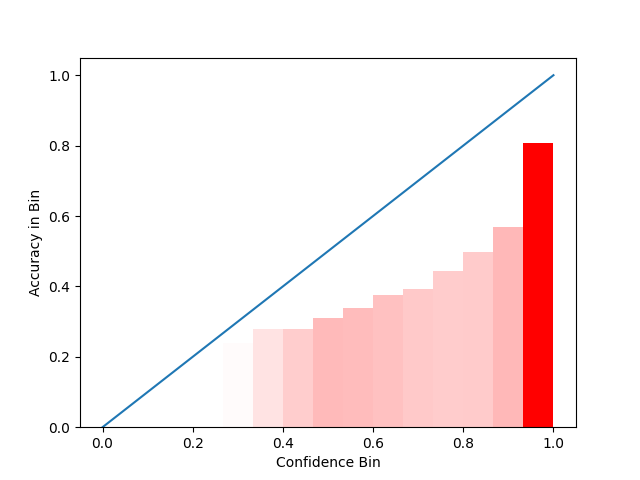}
        }
    \end{minipage}
    \hfill
    \begin{minipage}{0.45\textwidth}
        \centering
        \subfigure[Calibrated Llama-2-13b-Chat MMLU reliability diagram]{
            \includegraphics[width=\textwidth]{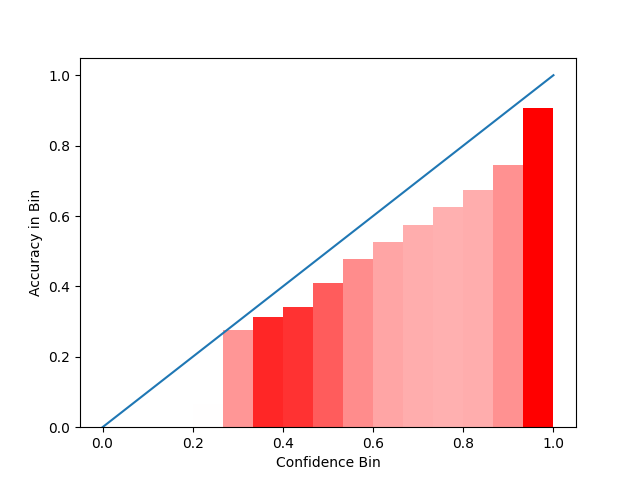}
        }
    \end{minipage}
    \vskip\baselineskip
    \begin{minipage}{0.45\textwidth}
        \centering
        \subfigure[Uncalibrated Llama-2-13b-Chat TriviaQA reliability diagram]{
            \includegraphics[width=\textwidth]{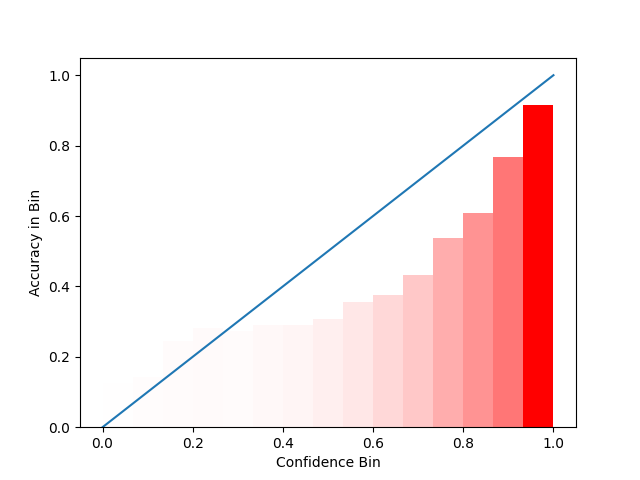}
        }
    \end{minipage}
    \hfill
    \begin{minipage}{0.45\textwidth}
        \centering
        \subfigure[Calibrated Llama-2-13b-Chat TriviaQA reliability diagram]{
            \includegraphics[width=\textwidth]{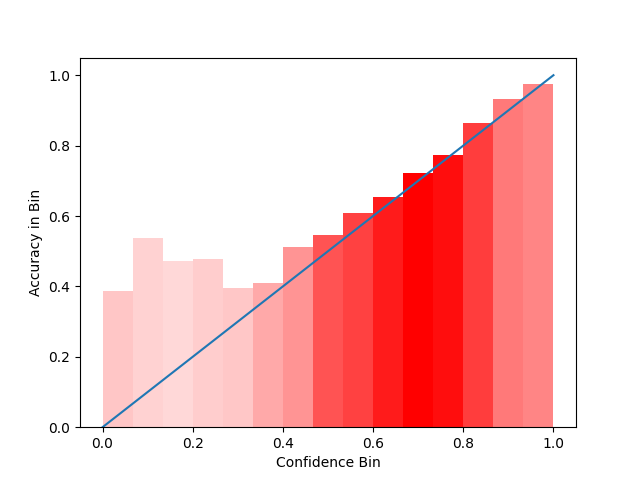}
        }
    \end{minipage}
    \vskip\baselineskip
    \begin{minipage}{0.45\textwidth}
        \centering
        \subfigure[Uncalibrated Llama-2-13b-Chat TruthfulQA reliability diagram]{
            \includegraphics[width=\textwidth]{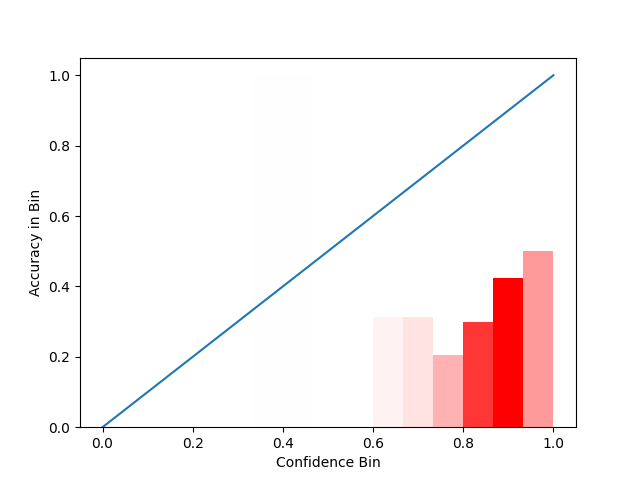}
        }
    \end{minipage}
    \hfill
    \begin{minipage}{0.45\textwidth}
        \centering
        \subfigure[Calibrated Llama-2-13b-Chat TruthfulQA reliability diagram]{
            \includegraphics[width=\textwidth]{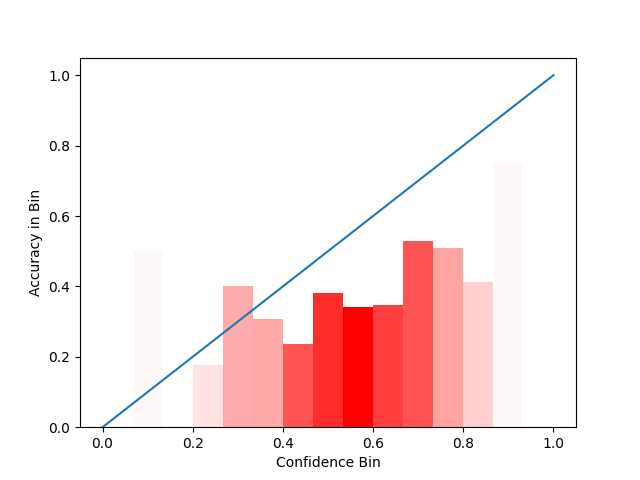}
        }
    \end{minipage}
    \caption{Llama-2-13b-Chat reliability diagrams.}
    \label{fig:llama13b_reliability_diagrams}
\end{figure*}

To better understand how our method changes the calibration of models, we show reliability diagrams for Llama-2-7b-Chat (Figure~\ref{fig:llama7b_reliability_diagrams}), Qwen-7b-Chat(Figure~\ref{fig:qwen_reliability_diagrams}), and Llama-2-13b-Chat(Figure~\ref{fig:llama13b_reliability_diagrams}). For each diagram we use 15 confidence bins, the same used in ECE evaluation. Additionally, we modify the transparency of bars based on the percentage of samples with confidence scores falling in each corresponding bin (more transparent indicating fewer samples). Additionally, confidence bins with no samples will not appear on the plot. A blue line showing perfect calibration is also drawn across each diagram for reference. The bar plots are plotted with the center of each bar corresponding to the confidence and accuracy value.

\end{document}